%% file: main.tex
\documentclass{jingdong}
\input{preamble}

\title{When Every Rollout Fails: \\ Splicing Learner-Conditioned Options into Long-Horizon Agent RL}
\author[1,2,3,*]{Haowen Hou}
\author[2,*]{Yicheng Liu}
\author[2]{Zhen Huang}
\author[3,2,\dagger]{Jiaqi Wang}

\affiliation[1]{Shanghai Jiao Tong University}
\affiliation[2]{Shanghai Innovation Institute}
\affiliation[3]{JD.com}
\contribution[*]{Equal contribution}
\contribution[\dagger]{Corresponding author}

\abstract{
Group-based reinforcement learning with verifiable rewards often yields all-fail groups on long-horizon agent tasks: every sampled trajectory receives zero reward, leaving group-relative methods without a reward-based policy gradient. A common remedy supplies complete solutions from a stronger model, but full trajectories are expensive and expose the learner to long off-policy histories. We introduce SPLICE, which constructs \textit{local} training targets from failed current-policy trajectories \textit{online}. For each all-fail group, a proposal model selects a splice point in a current-policy trajectory and generates several bounded options from that point. We estimate each option's value using repeated current-policy continuations, convert positive estimates into a return-weighted local target, and fit the current policy on the options. We evaluate SPLICE across two policy families on two long-horizon agent tasks, Software Engineering with SWE-bench Verified and Deep Research with BrowseComp-Plus. SPLICE outperforms GRPO and full-trajectory guidance baselines in all four settings, improving over GRPO by 2.8--9.8 percentage points. On initially failed SWE tasks, it raises GRPO's graduation rate by 10.8 percentage points. Further experiments show that SPLICE improves performance while reducing costly proposal-model computation.
}

\begin{document}
\maketitle

\section{Introduction}
\label{sec:introduction}

LLM agents solve long-horizon tasks through sequences of decisions and environment interactions~\citep{yao2023react,zhou2024webarena}. Reinforcement learning with verifiable rewards (RLVR) trains these agents using outcome feedback, such as whether a code change passes a test suite. Group Relative Policy Optimization (GRPO) assigns advantages by comparing rewards within a group of trajectories sampled for the same task~\citep{shao2024deepseekmath}. With binary outcome rewards, a group in which every trajectory fails has zero reward variation and provides no reward-based policy gradient. We refer to this case as an \emph{all-fail group}. Such a group leaves the policy without a successful trajectory to reinforce, even when some intermediate decisions could support a successful continuation.

External guidance offers a direct way to recover learning signals from all-fail groups by introducing successful behaviors that the current policy rarely generates. Existing methods for GRPO often derive such guidance from complete task-level solutions~\citep{yan2025luffy,ma2026relift,zhang2025bread}. This choice treats a complete solution as the source of guidance. For long-horizon agents, however, a failure may be localized: a short corrective segment can lead to a state from which the current policy can complete the remaining task.

Thus, using complete solutions in this long-horizon setting incurs both computational and distributional costs. A complete long-horizon trajectory may span tens or hundreds of interaction turns (e.g., Qwen3.5-9B rollouts on our SWE tasks average 83.7 agent turns). Repeatedly generating such trajectories with a strong proposal model requires expensive inference. The proposal model is requested even for decisions that the current policy can already make. Also, using long pre-collected trajectories as supervision exposes the learner to off-policy risk, increasing the mismatch between the training distribution and the current policy's behavior~\citep{luo2023forgetting,rajaraman2020limits,ross2011dagger}. These costs pose two design problems. The first is allocating proposal-model computation to bottlenecks encountered by current policy. The second is converting local proposals into a training target that reflects the current policy's ability to complete the task after each proposal.

We address both problems with SPLICE, which introduces failure-localized option splicing with current-policy continuation evaluation. For an all-fail group, a proposal model selects a splice point in one current-policy trajectory and generates several bounded options. We splice each option at this point, return control to the current policy, and sample repeated continuations to task completion. Their terminal outcomes estimate the option's current-policy continuation value, and we convert the positive values into a return-weighted local target. Only all-fail groups enter this splice branch, while other groups continue with standard GRPO. The bounded options save expensive proposal-model inference, while anchoring them in a current-policy history and supervising on the local target limits off-policy exposure.

We evaluate SPLICE on Software Engineering and Deep Research tasks. Across SWE-bench Verified and BrowseComp-Plus, it improves GRPO by 2.8--9.8 percentage points. On a fixed set of initially failed SWE tasks, it raises graduation rate from 8.1\% to 18.9\%. SPLICE uses fewer proposal-model rollout turns than full-trajectory guidance baselines while achieving better performance. Extensive experiments show that SPLICE can recover all-fail tasks without requiring the proposal model to generate or supervise complete trajectories.

Our contributions are as follows: \textbf{(1)} We introduce SPLICE, which formulates recovery from all-fail groups in long-horizon agent RL as learner-conditioned option splicing. This formulation allocates costly proposal-model computation only to splice points selected online. \textbf{(2)} We use current-policy continuation returns to define a target distribution over proposed options. By splicing bounded options into current-policy histories and fitting to this local distribution online, the update limits off-policy exposure relative to full-trajectory guidance. \textbf{(3)} We show across Software Engineering and Deep Research tasks that SPLICE improves performance and recovery of initially failed tasks while reducing proposal-model cost.

\section{Related Work}

\subsection{Agentic RL and All-Fail Groups}

Reinforcement learning has been extended from single-response reasoning to agents that act over multiple turns in stateful environments~\citep{wang2025ragen,yao2023react}. These settings often use sparse trajectory-level rewards from executable verifiers, which allow policy optimization over complete sequences of reasoning, actions, and environment observations. GRPO samples a group of trajectories for the same task and estimates each advantage from its return relative to the group, avoiding a learned value model~\citep{shao2024deepseekmath,dong2026rl}. The resulting update depends on reward variation among trajectories sampled from the same task.

Under binary outcome rewards, an all-fail group has zero reward for every trajectory, so its reward-relative advantages vanish. All-fail groups contain no positive trajectory to reinforce, making exploration on hard tasks the central problem. Existing methods reallocate sampling or filter zero-variance groups~\citep{yu2025dapo,zheng2025greso}, construct advantages from failed responses~\citep{le2025rlzvp}, or revise failures using gold answers~\citep{guo2026latentrevise}. External guidance offers another route by proposing behaviors that the current policy rarely samples, which we review below.

\subsection{External Guidance in RL}

One line of work conditions RL rollouts on precomputed hints or prefixes, while optimizing only the policy-generated continuation~\citep{zhang2025bread,setlur2026prefixrl,qu2026pope,ji2026actguide,yi2026pivotrl,nath2025guide}. Another line trains directly on complete external solutions through supervised, off-policy, or mixed-policy objectives~\citep{fu2025srft,zhang2025chord,ma2026relift,yan2025luffy}. Such guidance is often derived from expert solutions, while direct supervision requires complete expert trajectories. Generating these trajectories is expensive for long-horizon agents, and expert-induced histories can differ from those visited by the current policy, creating covariate shift and compounding errors~\citep{luo2023forgetting,rajaraman2020limits}. Concurrent work RSTG combines GRPO with on-policy distillation (OPD) on teacher solutions for all-fail groups~\citep{han2026rstg}, requiring aligned teacher and student token spaces. SPLICE instead queries a proposal model online for short options at selected splice points in current-policy trajectories. Learner-conditioned continuation evaluation and local supervision reduce proposal-model cost and limit off-policy exposure.

\section{Preliminaries}
\label{sec:preliminaries}

\textbf{Multi-turn agentic tasks.} We consider a task $x$ in a long-horizon environment. At agent turn $t$, the policy conditions on the interaction history $h_t=(x,o_1,a_1,\ldots,o_{t-1},a_{t-1},o_t)$ and samples an action $a_t\sim\pi_\theta(\cdot\mid h_t)$. The environment then returns an observation $o_{t+1}$, which may be terminal. A rollout produces a trajectory $\tau=(x,o_1,a_1,\ldots,o_T,a_T,o_{T+1})$ with binary outcome reward $R(\tau)\in\{0,1\}$.

\textbf{GRPO training.} At each update step, GRPO freezes a snapshot of the current policy $\pi_\theta$, denoted by $\pi_{\mathrm{old}}$, to collect trajectories while optimizing $\pi_\theta$. For each task, it samples a group of $G$ trajectories from $\pi_{\mathrm{old}}$ and estimates advantages without a learned value model~\citep{shao2024deepseekmath}. Let $\mathcal G(x)=\{\tau_i,R_i\}_{i=1}^{G}$ denote this group, where $R_i=R(\tau_i)$. The group-relative advantage is shared by all action tokens in trajectory $\tau_i$:
\begin{equation}
\widehat A_i=
\frac{R_i-\operatorname{mean}\!\left(\{R_j\}_{j=1}^{G}\right)}
{\operatorname{std}\!\left(\{R_j\}_{j=1}^{G}\right)+\epsilon_A}.
\label{eq:grpo_advantage}
\end{equation}
Flatten the policy-generated tokens across all turns of $\tau_i$ as $y_{i,1:N_i}$, and let $c_{i,n}$ denote the full context available to the policy when generating $y_{i,n}$. With the token-level importance ratio $r_{i,n}(\theta)=\pi_\theta(y_{i,n}\mid c_{i,n})/\pi_{\mathrm{old}}(y_{i,n}\mid c_{i,n})$, the reward-dependent GRPO loss is
\begin{equation}
\mathcal L_{\mathrm{GRPO}}(\theta)
=
-\frac{1}{G}\sum_{i=1}^{G}\frac{1}{N_i}\sum_{n=1}^{N_i}
\min\!\left(
r_{i,n}(\theta)\widehat A_i,
\operatorname{clip}\!\left(r_{i,n}(\theta),1-\epsilon_{\mathrm{low}},1+\epsilon_{\mathrm{high}}\right)\widehat A_i
\right).
\label{eq:grpo_objective}
\end{equation}
This trajectory-level use of GRPO assigns the terminal outcome to action tokens across all agent turns~\citep{wang2025ragen}.

If every trajectory has the same binary reward, Equation~\ref{eq:grpo_advantage} gives $\widehat A_i=0$ for all $i$, so the reward-dependent gradient in Equation~\ref{eq:grpo_objective} vanishes. An all-success group already contains successful behavior, whereas an all-fail group provides no successful trajectory to reinforce. SPLICE targets the latter case.

\section{Method}
\label{sec:splice}

\begin{figure}[!t]
\centering
\includegraphics[width=0.85\linewidth]{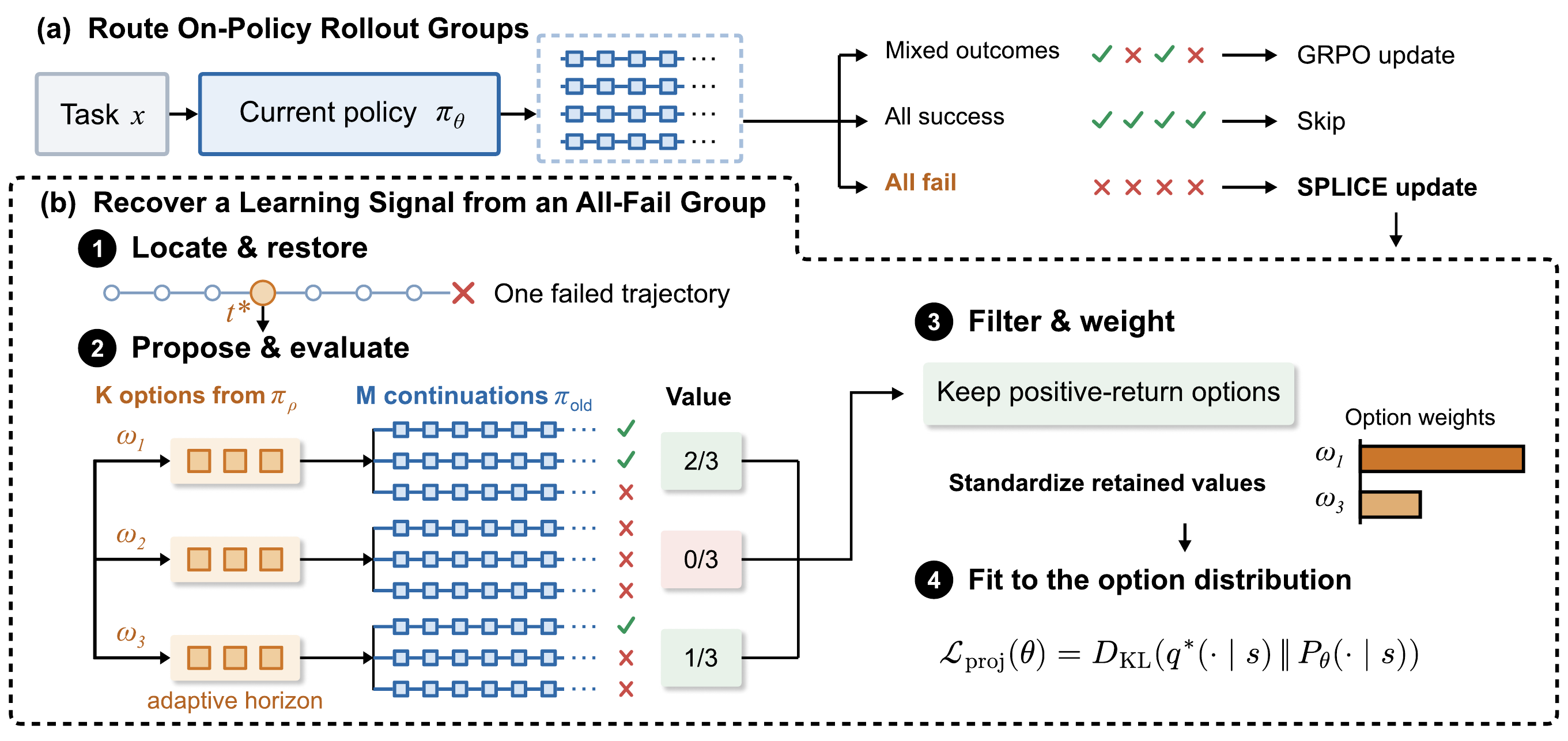}
\caption{Method overview. (a) We route each group of $G$ on-policy rollouts according to its binary outcomes. (b) The splice branch constructs a learner-conditioned target from short proposed options and current-policy continuations. We then fit the current policy to this target.}
\label{fig:method_overview}
\end{figure}

\textbf{Overview.} SPLICE supplements GRPO with a splice update for all-fail groups. Given an all-fail group, a proposal model $\pi_\rho$ selects an intermediate point in one failed trajectory and generates $K$ bounded options after resuming the interaction from that point. We splice each option at this point and evaluate it with $M$ continuations sampled from $\pi_{\mathrm{old}}$. We then retain options with positive estimated return and fit $\pi_\theta$ to their return-weighted distribution. Figure~\ref{fig:method_overview} summarizes this process.

\textbf{Update routing.} We first route each rollout group $\mathcal G$ according to its observed outcomes:
\begin{equation}
\operatorname{route}(\mathcal G)=
\begin{cases}
\textsc{GRPO}, & 0<\sum_{i=1}^{G}R_i<G,\\
\textsc{Skip}, & \sum_{i=1}^{G}R_i=G,\\
\textsc{Splice}, & \sum_{i=1}^{G}R_i=0.
\end{cases}
\label{eq:routing}
\end{equation}
Mixed-reward groups use Equation~\ref{eq:grpo_objective}, all-success groups are skipped, and all-fail groups enter the \textsc{Splice} branch described below.

\subsection{Option Construction and Learner-Conditioned Evaluation}
\label{sec:option_construction}

\textbf{Splice-point selection and branch reconstruction.} For each group $\mathcal G$ routed to \textsc{Splice}, we randomly select one failed trajectory $\tau^-$ from $\mathcal G$. The proposal model uses the task, full interaction history, and terminal verifier feedback to select $b_{t^*}$ among branchable points recorded every $C$ agent turns. A branchable point is a turn boundary from which the agent--environment interaction can be resumed. A stateful environment may restore a snapshot or reset and replay the recorded action prefix, while an environment with stateless tools may reuse the recorded history and issue subsequent requests afresh. We write $h_{t^*}$ for the history at $b_{t^*}$.

\textbf{Option proposal.} After resuming the selected branch, the proposal model generates a candidate set of options $\mathcal C_{\pi_\rho}(h_{t^*})=\{\omega_k\}_{k=1}^{K}$. Each option $\omega_k$ contains one or more agent turns generated by the proposal model. To bound the proposed segment, we limit each option to at most $H_b$ consecutive turns. For the current GRPO rollout batch $\mathcal B_b$, we set $H_b$ adaptively as
\begin{equation}
H_b=\max\!\left(1,\left\lceil\frac{\alpha}{|\mathcal B_b|}\sum_{\tau\in\mathcal B_b}T(\tau)\right\rceil\right),
\label{eq:option_horizon}
\end{equation}
where $T(\tau)$ is the number of agent turns in trajectory $\tau$ and $\alpha\in(0,1]$ controls the option length relative to the mean rollout length.

\textbf{Learner-conditioned evaluation.} Evaluating the proposed options serves two purposes: filtering incorrect options and measuring whether the current policy can use each option to complete the task. Directly training on unverified options can reinforce proposal errors.
Verifying them through proposal-model continuations addresses this issue, but they are costly and estimate option values under $\pi_\rho$. An option useful to $\pi_\rho$ may still be ineffective for $\pi_{\mathrm{old}}$. We therefore evaluate each nonterminal option through continuations from the current policy $\pi_{\mathrm{old}}$. Their returns estimate learner-conditioned option values and define a local target distribution over the retained options. Current-policy weighting and supervision on bounded option segments together limit off-policy exposure.

Concretely, we execute each option from the same splice point $b_{t^*}$. If $\omega_k$ completes the task during proposal, we immediately apply the task verifier and set $\widehat Q_k=1$ if it passes and $\widehat Q_k=0$ otherwise. If the task remains incomplete after $\omega_k$ ends, $\pi_{\mathrm{old}}$ generates $M$ independent continuations to task termination. Each resulting spliced trajectory $\tau_{k,j}$ starts with the interaction prefix reconstructed at $b_{t^*}$, followed by $\omega_k$ and the $j$th continuation. This construction replaces the failed suffix with a bounded proposed option followed by a current-policy continuation. The mean reward across continuations estimates the learner-conditioned value of option $\omega_k$:
\begin{equation}
\widehat Q_k=\frac{1}{M}\sum_{j=1}^{M}R(\tau_{k,j}).
\label{eq:option_value}
\end{equation}
We discard options with zero estimated return and skip the splice update if none remain. Let $\mathcal I=\{k:\widehat Q_k>0\}$ denote the retained candidate set.

\subsection{Return-Weighted Local Supervision}

For the splice update, we normalize each retained option value against the retained set:
\begin{equation}
\widetilde Q_k=
\frac{\widehat Q_k-\operatorname{mean}\!\left(\{\widehat Q_l\}_{l\in\mathcal I}\right)}
{\operatorname{std}\!\left(\{\widehat Q_l\}_{l\in\mathcal I}\right)+\epsilon_Q}.
\label{eq:q_normalization}
\end{equation}
These normalized values serve as candidate-relative advantages and require no value model. We convert them into a return-weighted candidate distribution:
\begin{equation}
q^*(\omega_k\mid h_{t^*})
=
\frac{\exp(\widetilde Q_k)}{\sum_{l\in\mathcal I}\exp(\widetilde Q_l)},
\label{eq:local_target}
\end{equation}
where $k\in\mathcal I$.

\begin{proposition}[Empirical dominance of return weighting]
\label{prop:return_weighting}
Fix a nonempty retained candidate set $\mathcal I$ and its empirical continuation returns $\{\widehat Q_k\}_{k\in\mathcal I}$. Let $u_k=1/|\mathcal I|$ be the uniform target over this set. The return-weighted target in Equation~\ref{eq:local_target} satisfies
\begin{equation}
\sum_{k\in\mathcal I}q^*(\omega_k\mid h_{t^*})\widehat Q_k
\geq
\frac{1}{|\mathcal I|}\sum_{k\in\mathcal I}\widehat Q_k.
\label{eq:empirical_dominance}
\end{equation}
The inequality is strict if the retained empirical returns are not all equal.
\end{proposition}

Proposition~\ref{prop:return_weighting} shows that, on the same retained candidates and observed continuation outcomes, return weighting assigns an empirical learner-conditioned value no lower than uniform supervision. This target-level result does not claim an improvement in the true return after a finite-sample policy update. Appendix~\ref{app:return_weighting_proof} provides the proof.

\textbf{Local policy update.} We use a sequence-level KL projection to motivate fitting the local target. For brevity, we write $h=h_{t^*}$ and $q_k^*=q^*(\omega_k\mid h)$, with $k$ indexing retained options.
Let $P_\theta(\cdot\mid h)$ denote the option distribution induced by $\pi_\theta$. The target $q^*$ is supported only on the retained options. The sequence-level projection is
\begin{equation}
\mathcal L_{\mathrm{proj}}(\theta)
=
D_{\mathrm{KL}}\!\left(q^*(\cdot\mid h)\,\Vert\,P_\theta(\cdot\mid h)\right).
\label{eq:local_projection}
\end{equation}
Holding $q^*$ fixed and expanding each option likelihood over its action tokens gives
\begin{equation}
\mathcal L_{\mathrm{proj}}(\theta)
=
-\sum_k q_k^*
\sum_{n=1}^{N_k}
\log\pi_\theta(y_{k,n}\mid c_{k,n})
+\mathrm{const}.
\label{eq:projection_nll}
\end{equation}
Here, $y_{k,1:N_k}$ denotes the $N_k$ action tokens within option $\omega_k$, and $c_{k,n}$ contains all earlier tokens in the trajectory. The constant collects terms independent of $\theta$. The sequence-level objective generally assigns a larger loss contribution to a longer option because it sums over its tokens. We therefore optimize the following length-normalized projection surrogate:
\begin{equation}
\mathcal L_{\mathrm{splice}}(\theta)
=
\sum_k q_k^*
\underbrace{\left[
-\frac{1}{N_k}
\sum_{n=1}^{N_k}
\log\pi_\theta(y_{k,n}\mid c_{k,n})
\right]}_{\text{mean token loss for option }\omega_k}.
\label{eq:splice_loss}
\end{equation}
This surrogate preserves the option-level target weights $q_k^*$ while making each option contribute through its mean token loss. It covers only proposal-model action tokens within retained options. Unlike Equation~\ref{eq:local_projection}, Equation~\ref{eq:splice_loss} is not an exact KL projection over variable-length option sequences.

\textbf{Training schedule.}
We keep GRPO and splice examples in separate mini-batches. For each GRPO rollout batch, we aggregate the eligible splice examples into a single splice mini-batch. We alternate training between GRPO and splice mini-batches. The proposal model and branch reconstruction are used only during training.

\section{Experiments}
\label{sec:experiments}

\subsection{Experimental Setup}

\textbf{Tasks and data.} For \textbf{Software Engineering (SWE)}, we combine subsets of SWE-Gym~\citep{pan2025swegym} and R2E-Gym~\citep{jain2025r2egym}, excluding tasks from repositories that overlap with SWE-bench. We evaluate resolve rate on SWE-bench Verified~\citep{jimenez2024swebench}. For \textbf{Deep Research}, we use a dataset curated from open-source datasets with a quality-control pipeline, which Appendix~\ref{app:deep-research-data} describes. We evaluate answer accuracy on BrowseComp-Plus~\citep{chen2026browsecompplus}, using Qwen3-Embedding-8B as the retriever and gpt-oss-20b as the judge model. To give all methods access to the same amount and distribution of training data, we sample a fixed subset of 3,200 tasks for both SWE and Deep Research. We report SWE and Deep Research performance in the main results, and further analyses use SWE.

\textbf{Models.} We study two policy families based on Qwen3-8B~\citep{qwen2025qwen3} and Qwen3.5-9B~\citep{qwen2026qwen35}. Our Qwen3-8B experiments use a shared cold-start checkpoint to adapt to the scaffold interfaces and tool-call formats, denoted Qwen3-8B-CS, whereas Qwen3.5-9B starts directly from its released checkpoint. Cold-start details are in Appendix~\ref{app:qwen3_cold_start}. Thinking is disabled for Qwen3-8B-CS for better performance, and is enabled for Qwen3.5-9B. We use Qwen3.5-122B-A10B as the proposal model for both policy families. We report main results for both families, and further analyses use Qwen3-8B-CS as the current policy.

\textbf{Agent scaffolds.} For SWE, we use OpenHands~\citep{wang2025openhands} as the scaffold. We keep the default system prompt and tools unchanged. For Deep Research, we customize the scaffold on Uni-Agent~\citep{ding2026uniagent}, providing three tools, \texttt{search}, \texttt{open}, and \texttt{find}.
For all training methods, the turn limit is 150 and the context limit is 128k tokens.

\textbf{Training configuration.} Unless varied in ablations, we use $K=3$ proposed options and $M=3$ current-policy continuations per option. The adaptive option horizon uses $\alpha=0.3$ in Equation~\ref{eq:option_horizon}. We record eligible splice points every $C=4$ agent turns. The default SWE implementation restores container snapshots, while Deep Research stores only the interaction history. Appendix~\ref{app:training_configuration} reports the hardware allocation and other hyperparameters.

\textbf{Reporting Convention.} We report mean $\pm$ sample standard deviation (SD) across three runs. We evaluate the final checkpoint from every training run.

\textbf{Comparison Setup.} We consider SPLICE, SFT, RL, SFT+RL, and ReLIFT, starting from the same initial policy and using the common 3,200-task subset for each task domain. This protocol controls the available task data rather than method-specific computation, which we compare separately in Section~\ref{sec:training_dynamics_cost}.
For \textbf{SFT}, the proposal model generates one full trajectory for every task in the common subset, and rejection sampling retains only successful trajectories for supervised fine-tuning.
\textbf{RL} applies GRPO directly to the initial policy.
\textbf{SFT+RL} initializes from the resulting SFT checkpoint and runs GRPO for the same number of RL steps as SPLICE.
We adapt \textbf{ReLIFT}~\citep{ma2026relift} to the online setting by requesting one full proposal-model trajectory for each task with an all-fail group and adding it to the buffer only upon verified success. Following the original method, ReLIFT triggers a fine-tuning step when the buffer contains at least 64 examples.

\subsection{Main Results}

\begin{table}[htbp]
\caption{Main results on SWE-bench Verified and BrowseComp-Plus. Qwen3-8B-CS is the cold-start checkpoint described in Appendix~\ref{app:qwen3_cold_start}.}
\label{tab:main_results}
\centering
\small
\setlength{\tabcolsep}{6.5pt}
\begin{tabular*}{0.774\linewidth}{@{\extracolsep{\fill}}lcccc}
\toprule
& \multicolumn{2}{c}{\textit{Qwen3-8B-CS}} & \multicolumn{2}{c}{\textit{Qwen3.5-9B}} \\
\cmidrule(lr){2-3}\cmidrule(lr){4-5}
Method & \shortstack{SWE-bench\\Verified} & \shortstack{BrowseComp-\\Plus} & \shortstack{SWE-bench\\Verified} & \shortstack{BrowseComp-\\Plus} \\
\midrule
Initial policy & 11.2 & 16.4 & 53.2 & 42.5 \\
SFT & $21.5{\scriptstyle\pm1.4}$ & $24.2{\scriptstyle\pm0.7}$ & $57.9{\scriptstyle\pm0.8}$ & $48.2{\scriptstyle\pm0.9}$ \\
RL & $15.9{\scriptstyle\pm0.7}$ & $20.2{\scriptstyle\pm0.9}$ & $58.4{\scriptstyle\pm0.4}$ & $46.2{\scriptstyle\pm0.4}$ \\
SFT+RL & $24.4{\scriptstyle\pm1.1}$ & $26.7{\scriptstyle\pm0.4}$ & $59.9{\scriptstyle\pm1.7}$ & $49.4{\scriptstyle\pm0.9}$ \\
ReLIFT & $19.7{\scriptstyle\pm0.6}$ & $23.5{\scriptstyle\pm0.6}$ & $59.1{\scriptstyle\pm1.0}$ & $47.6{\scriptstyle\pm1.9}$ \\
\addlinespace[2pt]
\rowcolor{familyrow}
SPLICE & $\mathbf{25.7{\scriptstyle\pm0.8}}$ & $\mathbf{29.3{\scriptstyle\pm0.7}}$ & $\mathbf{61.2{\scriptstyle\pm1.0}}$ & $\mathbf{50.7{\scriptstyle\pm0.3}}$ \\
\bottomrule
\end{tabular*}
\end{table}

Table~\ref{tab:main_results} first shows the limitation of applying RL directly to the initial policy. RL improves performance in all four settings, but the gains over the initial policy are limited to 3.7--5.2 percentage points. This modest improvement is consistent with the sparse-reward exploration bottleneck: GRPO can reinforce sampled successes but obtains no reward-based gradient from all-fail groups.

Providing successful trajectories before RL yields substantially stronger results. SFT+RL outperforms direct RL by 1.5--8.5 points across the four settings, while the RL stage improves its SFT initialization by a further 1.2--2.9 points. These results show that RL achieves better performance after SFT has exposed the policy to successful task trajectories.

Online external guidance offers an alternative way to supply successful behavior during RL. ReLIFT improves over direct RL by 0.7--3.8 points, showing that online guidance improves final performance in this comparison. SPLICE improves further over ReLIFT by 2.1--6.0 points and exceeds the higher-compute SFT+RL baseline by 1.3--2.6 points. It therefore achieves the highest performance in all four settings. These comparisons support the use of learner-conditioned local options rather than complete proposal trajectories for external guidance.

\subsection{Recovering Failed Tasks}
\label{sec:recovery}

We assess whether proposal guidance helps recover failed tasks and examine how training gains vary with task difficulty.

\textbf{Proposal Hit and Graduation Rates.} The \textbf{proposal hit rate} is the fraction of proposal attempts on SWE training tasks that yield at least one verified successful trajectory. The \textbf{graduation rate} is the fraction of initially failed SWE training tasks solved by the trained policy. We evaluate each trained method with one rollout per task, and a task \emph{graduates} if this rollout succeeds.

\begin{table}[htbp]
\centering
\small
\caption{Recovery and proposal-model cost on SWE training tasks. Proposal hit and graduation rates are percentages.}
\label{tab:proposal_recovery}
\setlength{\tabcolsep}{0pt}
\renewcommand{\arraystretch}{1.12}
\begin{tabular*}{0.64\linewidth}{@{\extracolsep{\fill}}lccc@{}}
\toprule
Method & \shortstack{Proposal hit} & \shortstack{Graduation} & $\pi_\rho$ Turns (k) \\
\midrule
RL & N/A & $8.1{\scriptstyle\pm1.1}$ & $0.0{\scriptstyle\pm0.0}$ \\
ReLIFT & $\mathbf{55.1{\scriptstyle\pm2.1}}$ & $15.9{\scriptstyle\pm1.0}$ & $113.4{\scriptstyle\pm5.9}$ \\
SPLICE & $30.7{\scriptstyle\pm0.5}$ & $\mathbf{18.9{\scriptstyle\pm0.3}}$ & $93.5{\scriptstyle\pm2.6}$ \\
\bottomrule
\end{tabular*}
\end{table}

As shown in Table~\ref{tab:proposal_recovery}, SPLICE obtains a proposal hit rate of 30.7\%, 24.4 percentage points below ReLIFT, while using 17.5\% fewer proposal-model turns. Despite the lower hit rate, SPLICE graduates 18.9\% of the initial-failure set, compared with 15.9\% for ReLIFT and 8.1\% for RL. The gains of 3.0 points over ReLIFT and 10.8 points over RL show that the splice updates yield stronger recovery in this experiment.

\textbf{Recovery Across Task Difficulty.} We compare training gains against standard GRPO across fixed Easy, Medium, and Hard subsets of the 3,200-task SWE training set; Appendix~\ref{app:task_difficulty_partition} defines the partition. Difficulty labeling and the evaluation in Table~\ref{tab:task_difficulty} use independent rollouts.

\begin{table}[htbp]
\centering
\small
\caption{Resolve rates (\%) by task difficulty on the SWE training set.}
\label{tab:task_difficulty}
\setlength{\tabcolsep}{0pt}
\renewcommand{\arraystretch}{1.12}
\begin{tabular*}{0.56\linewidth}{@{\extracolsep{\fill}}lccc@{}}
\toprule
Method & Easy & Medium & Hard \\
& ($n=342$) & ($n=1{,}087$) & ($n=1{,}771$) \\
\midrule
Initial & 99.4 & 0.5 & 0.2 \\
RL & $99.3{\scriptstyle\pm0.2}$ & $10.9{\scriptstyle\pm1.0}$ & $6.4{\scriptstyle\pm1.6}$ \\
SPLICE & $99.3{\scriptstyle\pm0.2}$ & $24.8{\scriptstyle\pm1.0}$ & $15.4{\scriptstyle\pm0.4}$ \\
\bottomrule
\end{tabular*}
\end{table}

Table~\ref{tab:task_difficulty} shows that the largest improvement occurs on Medium tasks. Easy-task performance is already near saturation. On Medium tasks, SPLICE gains 24.3 points, exceeding RL's gain by 13.9 points. This gap suggests that local option splicing is particularly useful at intermediate difficulty, where standard RL makes limited progress but targeted corrections can still enable recovery. On Hard tasks, SPLICE gains 15.2 points compared with 6.2 points for RL.

\FloatBarrier

\subsection{Training Dynamics and Cost}
\label{sec:training_dynamics_cost}

We analyze training along three complementary axes: resolve-rate dynamics over training updates, proposal-model rollout turns, and total wall-clock time. Proposal-model turns isolate generation by the expensive guidance model, whereas wall-clock time measures end-to-end training cost.

\begin{figure}[ht]
\centering
\includegraphics[width=0.90\linewidth]{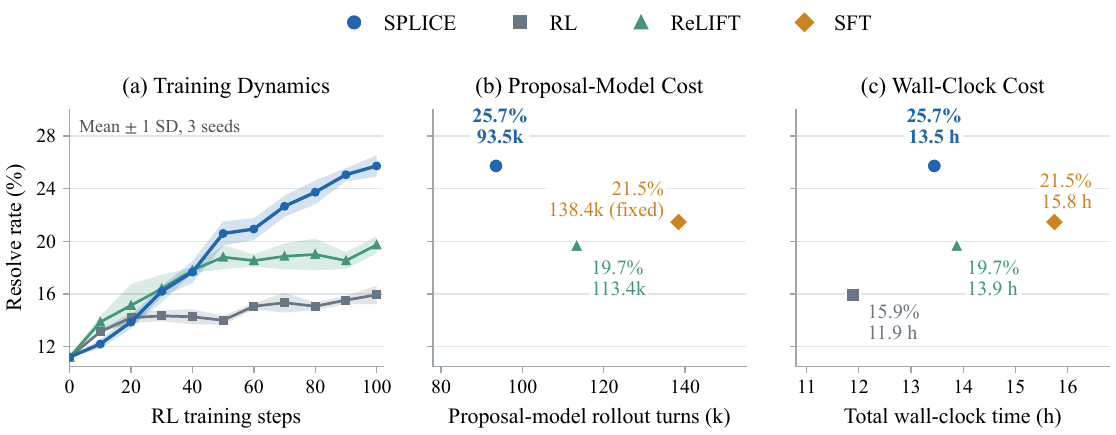}
\caption{Training dynamics and cost. (a) Resolve rate every 10 RL steps. (b,c) Resolve rate versus proposal-model turns and total wall-clock time, respectively. Higher and further left is better.}
\label{fig:training_dynamics_cost}
\end{figure}

\textbf{Training Dynamics.} Figure~\ref{fig:training_dynamics_cost}(a) compares how resolve rate changes during training. RL improves quickly but later fluctuates at a relatively low level. ReLIFT improves over a longer horizon, while SPLICE shows no early advantage but later surpasses both baselines. Thus, external guidance supports continued improvement, with SPLICE achieving the strongest late-stage performance.

\textbf{Proposal-Model and Wall-Clock Cost.} Figure~\ref{fig:training_dynamics_cost}(b,c) compares proposal-model generation and end-to-end training time. SFT collects a complete proposal-model trajectory for every task, giving it the highest proposal-model cost and longest runtime, whereas RL uses no proposal model and finishes fastest. ReLIFT and SPLICE spend additional time on online external guidance but outperform RL. For both methods, we asynchronously generate rollout batches for RL and supervised updates with $\text{staleness}=2$ and observe comparable resolve rates to synchronous training; Appendix~\ref{app:training_configuration} provides configuration details. SPLICE requests only bounded proposal options rather than complete trajectories; despite the additional student continuations used to evaluate them, it requires less proposal-model computation and wall-clock time than ReLIFT and SFT while achieving the highest resolve rate.

\FloatBarrier

\subsection{Mechanism and Ablation Studies}
\label{sec:mechanism_ablations}

We test four design choices: how continuation outcomes define the local target, which policy generates the continuations, how much of each proposal-generated branch enters the loss, and how long each option should be. All variants are trained under the same setting, except for the changes stated in each comparison. \textbf{Resolve} measures SWE-bench Verified performance; \textbf{Graduation} and \textbf{Hit} are training-set metrics defined in Section~\ref{sec:recovery}. Appendix~\ref{app:splice_point_selection} studies strategies for selecting splice points.

\textbf{Continuation Evaluation and Target Construction.} We compare SPLICE's return-weighted target with three alternatives. \textbf{Single option} directly supervises one generated option without further verification. \textbf{Hard top-$\widehat Q$} retains $K=M=3$ continuation evaluation but places all target mass on the option attaining the largest positive $\widehat Q_k$. We randomly select one if multiple options have the same $\widehat Q_k$. \textbf{Uniform positive} uses the same continuation outcomes but assigns equal weight to every option with $\widehat Q_k>0$. SPLICE improves Resolve by 4.5 points and Graduation by 6.6 points over single-option supervision. More directly, it exceeds hard top-$\widehat Q$ by 1.4 Resolve points and 1.2 Graduation points under matched proposal and continuation budgets. This result supports retaining a soft distribution over verified alternatives rather than collapsing the target onto the best. Relative to uniform positive, return-dependent weighting adds 2.0 Resolve points and 1.8 Graduation points, showing that the relative continuation returns provide useful information beyond binary outcome filtering.

\begin{table}[htbp]
\centering
\small
\caption{Mechanism ablations. Bold text marks the default configuration.}
\label{tab:mechanism_ablations}
\setlength{\tabcolsep}{0pt}
\renewcommand{\arraystretch}{1.10}

\begin{tabular*}{0.70\linewidth}{@{}p{0.34\linewidth}@{\extracolsep{\fill}}cc@{}}
\toprule
\multicolumn{3}{l}{\textbf{Continuation evaluation and option weighting}} \\
Setting & Resolve & Graduation \\
\midrule
\textbf{SPLICE} & $\mathbf{25.7{\scriptstyle\pm0.8}}$ & $\mathbf{18.9{\scriptstyle\pm0.3}}$ \\
Single option & $21.2{\scriptstyle\pm0.4}$ & $12.3{\scriptstyle\pm1.1}$ \\
Hard top-$\widehat Q$ & $24.3{\scriptstyle\pm0.6}$ & $17.7{\scriptstyle\pm0.2}$ \\
Uniform positive & $23.7{\scriptstyle\pm0.2}$ & $17.1{\scriptstyle\pm1.1}$ \\
\bottomrule
\end{tabular*}
\vspace{3pt}
\begin{tabular*}{0.70\linewidth}{@{\extracolsep{\fill}}llccc@{}}
\multicolumn{5}{l}{\textbf{Continuation policy and supervision scope}} \\
Policy & Scope & Resolve & Graduation & $\pi_\rho$ Turns (k) \\
\midrule
\textbf{Current} & \textbf{Option} & $\mathbf{25.7{\scriptstyle\pm0.8}}$ & $\mathbf{18.9{\scriptstyle\pm0.3}}$ & $\mathbf{93.5{\scriptstyle\pm2.6}}$ \\
Proposal & Option & $26.2{\scriptstyle\pm1.6}$ & $19.4{\scriptstyle\pm1.3}$ & $372.5{\scriptstyle\pm12.0}$ \\
Proposal & Suffix & $22.3{\scriptstyle\pm1.5}$ & $15.4{\scriptstyle\pm1.4}$ & $365.9{\scriptstyle\pm18.3}$ \\
\bottomrule
\end{tabular*}
\vspace{3pt}
\begin{tabular*}{0.70\linewidth}{@{\extracolsep{\fill}}lcccc@{}}
\multicolumn{5}{l}{\textbf{Adaptive option horizon}} \\
$\alpha$ & Resolve & Hit & Graduation & $\pi_\rho$ Turns (k) \\
\midrule
0.1 & $18.2{\scriptstyle\pm1.1}$ & $11.3{\scriptstyle\pm1.6}$ & $10.7{\scriptstyle\pm1.9}$ & $33.7{\scriptstyle\pm0.9}$ \\
\textbf{0.3} & $\mathbf{25.7{\scriptstyle\pm0.8}}$ & $\mathbf{30.7{\scriptstyle\pm0.5}}$ & $\mathbf{18.9{\scriptstyle\pm0.3}}$ & $\mathbf{93.5{\scriptstyle\pm2.6}}$ \\
0.5 & $24.3{\scriptstyle\pm1.3}$ & $33.0{\scriptstyle\pm0.7}$ & $17.1{\scriptstyle\pm1.1}$ & $157.0{\scriptstyle\pm6.9}$ \\
\bottomrule
\end{tabular*}
\end{table}

\textbf{Continuation Policy and Supervision Scope.} We first replace current-policy continuations with \textbf{proposal-model continuations} while retaining supervision only on the bounded options. This change improves Resolve from 25.7\% to 26.2\% and Graduation from 18.9\% to 19.4\%, but raises proposal-model rollout turns from 93.5k to 372.5k, or $4.0\times$ as many turns. The \textbf{full-suffix variant} also supervises the action tokens in successful proposal-generated continuations. To control the contribution of sequence length, we average the token loss within each branch and then average successful branches within each option before applying its target weight. Full-suffix supervision lowers both Resolve and Graduation while using 365.9k proposal-model turns. The two variants therefore use comparable proposal-model computation, but extending supervision to the successful suffix degrades performance. This may be attributed to greater off-policy exposure along proposal-induced histories and accumulated covariate shift in long trajectories.

\textbf{Adaptive Option Horizon.} We vary the fraction in Equation~\ref{eq:option_horizon} over $\alpha\in\{0.1,0.3,0.5\}$, with $\alpha=0.3$ as the reference. \textbf{Reducing $\alpha$ to 0.1} cuts proposal-model turns from 93.5k to 33.7k, a 64.0\% reduction, but lowers Hit, Resolve and Graduation. This pattern suggests that very short options are less likely to reach a state from which the current policy can complete the task. \textbf{Increasing $\alpha$ to 0.5} raises Hit by 2.3 points and proposal-model turns from 93.5k to 157.0k, a 67.9\% increase. Resolve and Graduation nevertheless slightly decrease. Longer teacher options therefore do not yield a better proposal-free policy in this comparison.

\FloatBarrier

\begin{figure}[htbp]
\centering
\includegraphics[width=0.80\linewidth]{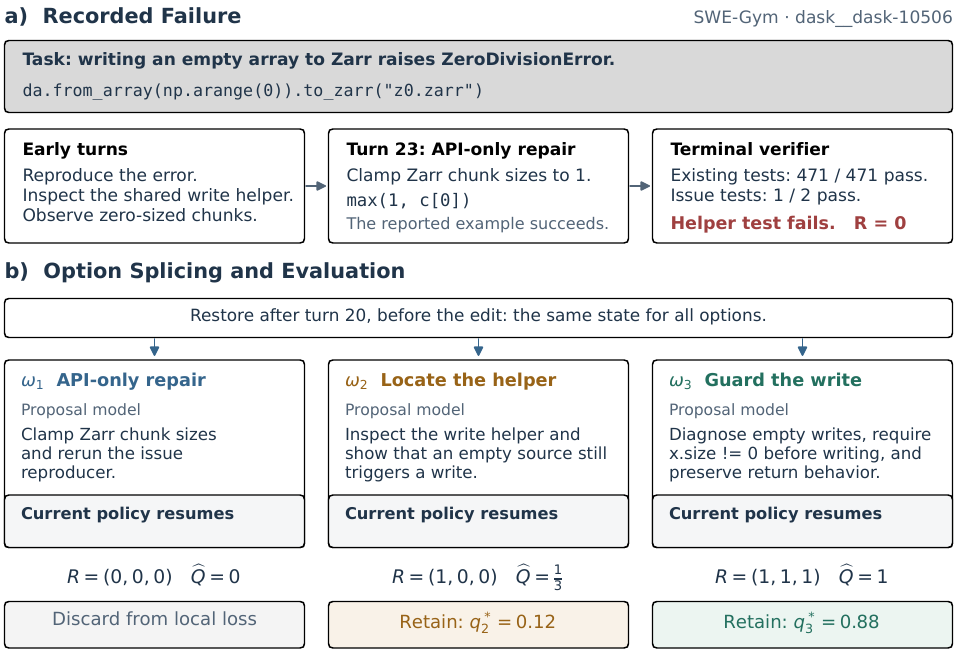}
\caption{A case sampled from training with summarized trajectory steps. (a) Recorded actions and verifier feedback from a failed rollout. (b) Options from the proposal model, followed by current-policy continuations.}
\label{fig:case_study}
\end{figure}

\subsection{Case Study: Redirecting an Incomplete Repair}
\label{sec:case_study}

We illustrate the splice update with SWE-Gym instance \texttt{dask\_\_dask-10506} in Figure~\ref{fig:case_study}, which concerns a division-by-zero error when writing an empty array to Zarr. The raw steps are summarized for clarity.

\textbf{An Incomplete Repair.}
We randomly select one trajectory in this all-fail group. It passes 471 existing regression tests and one of the two issue-specific tests, but fails \texttt{test\_load\_store\_chunk}. In this trajectory, the agent reproduces the exception, inspects the shared write helper, and then changes Zarr chunk sizes from zero to one. This change fixes the reported example, but the helper still attempts to write empty blocks when called directly.

\textbf{Option Splicing and Verification.}
The proposal model selects the branchable point after 20 completed turns, before the first production edit. All three options begin from this point. After each option, the current policy produces three independent continuations.
The returns are $(0,0,0)$, $(1,0,0)$, and $(1,1,1)$ for $\omega_1$, $\omega_2$, and $\omega_3$, respectively. Thus, the diagnostic option sometimes leads the current policy to the complete repair. These give $\widehat Q=(0,\frac13,1)$: the API-only option is discarded, while the diagnostic and guarded-write options remain. Equations~\ref{eq:q_normalization}--\ref{eq:local_target} assign the retained options weights of approximately $0.12$ and $0.88$. The resulting target favors the guarded helper repair while retaining some probability on the diagnostic behavior that may also support a successful completion. This example illustrates how short, outcome-verified alternatives can extract a local training target from an all-fail group.

\FloatBarrier

\section{Limitations and Conclusion}

\textbf{Limitations.} SPLICE introduces current-policy rollout computation because each option is evaluated through multiple continuations. Our asynchronous pipeline adds limited wall-clock time in our experiments, but the cost may be larger in environments with expensive execution. Also, estimating each option value from three continuations introduces sampling error. Increasing its accuracy requires more current-policy continuations or a stronger model, creating a trade-off between estimation quality and inference cost.

\textbf{Conclusion.} We introduced SPLICE, a training method that recovers learning signals from all-fail rollout groups in long-horizon agent RL. It splices bounded proposal-model options into failed current-policy trajectories online. We evaluate each option through current-policy continuations and train on the resulting return-weighted local target. Replacing complete proposal trajectories with local options online reduces proposal-model cost and limits off-policy exposure and covariate shift. Across two policy families and two agent tasks, SPLICE outperforms GRPO and full-trajectory guidance, with particularly strong recovery on initially failed tasks.

\FloatBarrier
\bibliographystyle{ieeenat_fullname}
\bibliography{main}


\clearpage
\appendix
\section*{Appendix}

\section{Proof of Proposition~\ref{prop:return_weighting}}
\label{app:return_weighting_proof}

\begin{proof}
Let $m=|\mathcal I|$ and $u_k=1/m$. Define $d=\operatorname{std}(\{\widehat Q_l\}_{l\in\mathcal I})+\epsilon_Q>0$ and $\overline Q=m^{-1}\sum_{k\in\mathcal I}\widehat Q_k$. The normalized return then satisfies $\widetilde Q_k=(\widehat Q_k-\overline Q)/d$. Consider the KL-regularized objective
\begin{equation}
F(q)=\sum_{k\in\mathcal I}q_k\widetilde Q_k-D_{\mathrm{KL}}(q\Vert u),
\qquad q\in\Delta(\mathcal I).
\label{eq:regularized_candidate_objective}
\end{equation}
This objective is strictly concave in $q$. Its Lagrangian stationarity condition gives
\begin{equation}
q_k\propto u_k\exp(\widetilde Q_k).
\end{equation}
Because $u$ is uniform, the unique maximizer is exactly the softmax target $q^*$ in Equation~\ref{eq:local_target}. Since $u$ is feasible, optimality gives
\begin{align}
F(q^*)&\geq F(u),\\
\sum_{k\in\mathcal I}q_k^*\widetilde Q_k
&\geq
\sum_{k\in\mathcal I}u_k\widetilde Q_k
+D_{\mathrm{KL}}(q^*\Vert u).
\label{eq:normalized_return_dominance}
\end{align}
The affine relation $\widehat Q_k=\overline Q+d\widetilde Q_k$ preserves comparisons between expectations under distributions on the fixed set $\mathcal I$. Dropping the nonnegative KL term in Equation~\ref{eq:normalized_return_dominance} therefore yields Equation~\ref{eq:empirical_dominance}.

If the retained empirical returns are not all equal, then $\widetilde Q$ is nonconstant and its softmax is nonuniform. Hence $q^*\neq u$ and $D_{\mathrm{KL}}(q^*\Vert u)>0$. In this case, Equations~\ref{eq:normalized_return_dominance} and~\ref{eq:empirical_dominance} are strict. If all retained empirical returns are equal, then $q^*=u$ and equality holds.
\end{proof}

\section{Training Data}
\label{app:training_data}

\subsection{Software Engineering}

We construct the SWE training pool from subsets of SWE-Gym~\citep{pan2025swegym} and R2E-Gym~\citep{jain2025r2egym}. SWE-Gym collects real-world issue-resolution tasks from pull requests in Python repositories and pairs each task with an executable environment and expert-validated tests~\citep{pan2025swegym}. It contains 2,438 tasks, while we omit 31 tasks because their container images are unavailable. R2E-Gym procedurally constructs executable tasks from repository commits through back-translation and test collection or generation~\citep{jain2025r2egym}. We use its subset of 4,578 tasks from repositories that do not overlap with SWE-bench.

\subsection{Deep Research Dataset Construction}
\label{app:deep-research-data}

We draw candidate tasks from five public releases, which are OpenResearcher~\citep{li2026openresearcher}, DeepForge from OffSeeker~\citep{zhou2026offseeker}, OpenSeeker~\citep{du2026openseeker}, REDSearcher~\citep{chu2026redsearcher} and public Deep Research SFT datasets released by Kyle Montgomery~\citep{montgomery2026deepresearch}. Some releases provide complete interaction traces in addition to question--answer pairs. We extract only the query and reference answer.

We canonicalize each query with Unicode NFKC normalization, case folding, and whitespace collapsing. We then deduplicate the normalized queries jointly across all five sources, obtaining 75,372 unique candidates.

Collection dates and retrieval systems differ across these releases, so a reference answer may be ambiguous, outdated, or unsupported in our web environment. We audit every candidate with DeepSeek-V4-Flash~\citep{deepseekai2026deepseekv4} under a unified environment. The model can call \texttt{search}, \texttt{open}, and \texttt{find}, and Serper provides web search and reading services.\footnote{\url{https://serper.dev/}} We accept a task only when the audit answer agrees with its reference and the current search results provide supporting evidence. This common audit removes ill-posed questions, stale answers, and tasks whose evidence is no longer reachable. The final pool contains 16,855 Deep Research training tasks.

\section{Cold-start Training for Qwen3-8B}
\label{app:qwen3_cold_start}

We apply cold start only to Qwen3-8B. This stage adapts the released checkpoint to the agent scaffolds and their tool-call formats. Qwen3.5-9B starts directly from its released checkpoint.

We sample 500 SWE tasks and 500 Deep Research tasks from the data pool described in Appendix~\ref{app:training_data}. Qwen3.5-122B-A10B generates candidate trajectories, and rejection sampling retains trajectories that pass the corresponding verifier. We mix the retained SWE and Deep Research trajectories into one set and fine-tune the released Qwen3-8B checkpoint to obtain Qwen3-8B-CS.

Table~\ref{tab:qwen3_cold_start_results} separates the effect of cold start from subsequent training. Under the evaluation setup used in the main experiments, the released Qwen3-8B checkpoint obtains 8.6\% on SWE-bench Verified and 12.7\% on BrowseComp-Plus. Qwen3-8B-CS obtains 11.2\% and 16.4\%, respectively. All Qwen3-8B methods in Table~\ref{tab:main_results} start from this same Qwen3-8B-CS checkpoint.

\begin{table}[htbp]
\caption{Qwen3-8B performance before method-specific training.}
\label{tab:qwen3_cold_start_results}
\centering
\small
\setlength{\tabcolsep}{9pt}
\begin{tabular}{lcc}
\toprule
Initial checkpoint & SWE-bench Verified (\%) & BrowseComp-Plus (\%) \\
\midrule
Qwen3-8B (released) & 8.6 & 12.7 \\
Qwen3-8B-CS & 11.2 & 16.4 \\
\bottomrule
\end{tabular}
\end{table}

\section{Training Configuration}
\label{app:training_configuration}

We train on four compute nodes, each containing 8 NVIDIA H200 GPUs. For the RL baseline, all four nodes run the student policy, with the trainer and rollout actor colocated on the same GPUs. For SPLICE and ReLIFT, three nodes use this colocated student setup, while the remaining node serves the proposal model. Training is asynchronous with a maximum rollout staleness of 2. Each rollout batch contains 32 groups, with $G=8$ trajectories per group, and the actor learning rate is $2\times10^{-6}$. For the SFT baseline, we train for three epochs with a learning rate of $1\times10^{-5}$.

\section{Task-Difficulty Partition}
\label{app:task_difficulty_partition}

We construct fixed difficulty subsets from the 3,200-task SWE training set before training. We run Qwen3-8B-CS and Qwen3-Coder-30B-A3B-Instruct once on each task and assign labels from their verified outcomes. A task is \emph{Easy} if Qwen3-8B-CS succeeds, regardless of the Qwen3-Coder-30B-A3B-Instruct outcome. Among the remaining tasks, those solved by Qwen3-Coder-30B-A3B-Instruct are \emph{Medium}, and those on which both models fail are \emph{Hard}. The labeling runs yield success rates of 10.7\% for Qwen3-8B-CS and 43.6\% for Qwen3-Coder-30B-A3B-Instruct, with 308 tasks solved by both models. The resulting partition contains 342 Easy, 1,087 Medium, and 1,771 Hard tasks.

\section{Splice-Point Selection}
\label{app:splice_point_selection}

We compare three rules for choosing the splice point using the setup in Section~\ref{sec:mechanism_ablations}.
Each rule samples one failed trajectory uniformly from the all-fail group and restricts selection to branchable points recorded every $C=4$ turns. In this SWE experiment, the default implementation resumes these points through container snapshots.
\emph{Random} selects an eligible splice point uniformly.
\emph{Highest entropy} selects the eligible splice point whose recorded action has the largest mean token entropy under $\pi_{\mathrm{old}}$.
For the $N_t$ action tokens at turn $t$, we compute
\begin{equation}
\mathcal H_t=-\frac{1}{N_t}\sum_{n=1}^{N_t}\sum_{v\in\mathcal V}\pi_{\mathrm{old}}(v\mid c_{t,n})\log\pi_{\mathrm{old}}(v\mid c_{t,n}),
\label{eq:turn_entropy}
\end{equation}
where $\mathcal V$ is the policy vocabulary and $c_{t,n}$ is the context before token $n$ in that turn.
Averaging over tokens avoids ranking turns by action length; ties select the earliest eligible turn.
\emph{Proposal model} uses the selector described in Section~\ref{sec:option_construction}.

\begin{table}[htbp]
\caption{Splice-point selection ablation. Resolve is the SWE-bench Verified resolve rate, and Graduation is the success rate on initially failed SWE training tasks. Both are percentages.}
\label{tab:splice_point_selection}
\centering
\small
\setlength{\tabcolsep}{7pt}
\begin{tabular*}{0.5525\linewidth}{@{\extracolsep{\fill}}lcc}
\toprule
Splice-point selection & Resolve & Graduation \\
\midrule
Random & $24.4{\scriptstyle\pm0.4}$ & $17.8{\scriptstyle\pm2.3}$ \\
Highest entropy & $23.9{\scriptstyle\pm1.5}$ & $17.2{\scriptstyle\pm0.8}$ \\
Proposal model & $\mathbf{25.7{\scriptstyle\pm0.8}}$ & $\mathbf{18.9{\scriptstyle\pm0.3}}$ \\
\bottomrule
\end{tabular*}
\end{table}

Highest-entropy selection lowers Resolve by 0.5 points and Graduation by 0.6 points relative to random selection (Table~\ref{tab:splice_point_selection}), indicating that local policy uncertainty alone does not identify better splice points.
Proposal-model selection improves Resolve by 1.3 points and Graduation by 1.1 points over random selection; relative to highest-entropy selection, the gains are 1.8 and 1.7 points, respectively.
These results suggest that conditioning on the full failed trajectory and terminal feedback provides a more useful selection signal than either random choice or local token entropy.

\end{document}

%% file: preamble.tex
\usepackage{amsmath,amssymb,amsthm}
\usepackage{enumitem}
\usepackage{colortbl}

\definecolor{familyrow}{RGB}{232,240,248}
\newtheorem{proposition}{Proposition}

\crefname{figure}{Fig.}{Figs.}
\Crefname{figure}{Fig.}{Figs.}
\crefname{table}{Table}{Tables}
\Crefname{table}{Table}{Tables}